\documentclass[oneside,ngerman]{article}

\usepackage[utf8]{inputenc}
\usepackage{graphicx}
\usepackage{multirow}
\usepackage{hyperref}
\usepackage{float}
\usepackage[T1]{fontenc}
\usepackage{amsmath,amssymb}
\usepackage{caption}
\usepackage{color}
\usepackage{wrapfig}
\usepackage[font=small]{subcaption}
\usepackage{float}
\usepackage{bbding}
\usepackage{multirow}
\let\vec\mathbf
\newcommand{\ds}{\displaystyle}

\newcommand{\vecp}{\vec{p}}
\newcommand{\mat}[1]{\text{\textbf{#1}}}
\newcommand{\vecpConst}{\vec{p}^{init}_{const}}
\newcommand{\vecpRnd}{\vec{p}^{init}_{rnd}}

\title{\vspace{-2.5cm}{\bf Collocation Polynomial Neural Forms and 
Domain Fragmentation for solving Initial Value Problems}}
\author{Toni Schneidereit and Michael Breuß \\ 
\\[1ex]
\small Applied Mathematics Group \\ 
\small Brandenburg University of Technology Cottbus-Senftenberg \\
\small Platz der Deutschen Einheit 1, 03046 Cottbus, Germany \\
\small \{Toni.Schneidereit,breuss\}@b-tu.de
}

\date{\small \today} 

\begin{document}

\maketitle

\begin{abstract}

Several neural network approaches for solving differential equations employ trial solutions with a feedforward neural network. There are different means to incorporate the trial solution in the construction, for instance one may include them directly in the cost function. Used within the corresponding neural network, the trial solutions define the so-called neural form. Such neural forms represent general, flexible tools by which one may solve various differential equations. 

In this article we consider time-dependent initial value problems, which require to set up the neural form framework adequately. The neural forms presented up to now in the literature for such a setting can be considered as first order polynomials. In this work we propose to extend the polynomial order of the neural forms. The novel collocation-type
construction includes several feedforward neural networks, one for 
each order. Additionally, we propose the fragmentation of the computational domain into subdomains. The neural forms are solved on each subdomain, whereas the interfacing grid points overlap in order to provide initial values over the whole fragmentation. 

We illustrate in experiments that the combination of collocation neural forms of higher order and the domain fragmentation allows to solve initial value problems over large domains with high accuracy and reliability.
\end{abstract}
\begin{center}
\vspace{0.875cm}
\end{center}

\section{Introduction}

Over the last decades several neural network approaches for solving differential equations have been developed \cite{Yadav2015IntroNN,Maede1994lin,Dissanayake1994approt}. The application and extension of these approaches is a topic of recent research, including work on different network architectures like Legendre \cite{mall2016legendre} and polynomial neural networks \cite{Zjavka2016PNN} as well as computational studies \cite{Famelis2020study,Schneidereit2020ODEANN}. \par 
One of the early proposed methods \cite{lagaris1998artificial} introduced a trial solution (TS) in order to define a cost function using one feedforward neural network. The TS is supposed to satisfy given initial or boundary values by construction. It is also referred to as neural form (NF) in this context \cite{lagaris1998artificial,Lagari2020Neural} which we will adopt from here on. Let us note that such NFs represent a general tool that enable to solve ordinary ordinary differential equations (ODEs), partial differential equations (PDEs) and systems of ODEs/PDEs alike. 
We will refer here to this approach as the trial solution method (TSM). Later, the initial method from \cite{lagaris1998artificial} has been extended by a NF with two feedforward neural networks, which allows to deal with boundary value problems for irregular boundaries \cite{lagaris2000boundary} and yields broader possibilities for constructing the TS \cite{Lagari2020Neural}. In the latter context, let us also mention \cite{TSoulos2009ConstTS} where an algorithm is proposed in order to create a TS based on grammatical evolution. Focusing on initial value problems (IVPs), one approach employs to learn solution bundles \cite{Flamant2020bundles}, making the trained neural forms reusable for enquired initial values. \par
A technique related to TSM that avoids the explicit construction of trial solutions has been proposed in \cite{piscopo2019solving}. The given initial or boundary values from the underlying differential equation are included in the cost function as additional terms, so that the NF can be set to equal the neural network output. We will refer to this approach as modified trial solution method (mTSM). \par
The fact that the neural network output computation resembles a linear combination of basis functions leads to a network architecture as presented in \cite{Rudd2015CINT} (for PDEs). In that work one hidden layer incorporates two sets of activation functions, one of which is supposed to satisfy the PDE and the second dealing with boundary conditions. The basis function coefficients are set to be the connecting weights from the hidden layer to the output neuron, and the sum over all basis functions and coefficients makes up the NF. \par 

Motivated by the construction principle of collocation methods in numerical analysis, we propose in this paper a novel extension of the NF approach. Our neural form extension is based on the observation, that the NF using one feedforward neural network as employed in \cite{lagaris1998artificial} may be interpreted as a first order collocation polynomial. We propose to extend the corresponding polynomial order of the neural form. The novel construction includes several feedforward neural networks, one for each order. Compared to a collocation method from standard numerics, the networks take on the role of coefficients in the collocation polynomial expansion. \par

Furthermore, we aim to approximate initial value problems on fairly large domains. Therefore, and based on the NF structures, we also propose a fragmentation of the computational domain into subdomains. In each subdomain, we solve the initial value problem with a collocation neural form. This is done proceeding in time from one domain fragment to the adjacent subdomain. The interfacing grid points in any subdomain provide the initial value for the next subdomain. On a first glance one may think of similarities to domain decomposition methods for PDEs in numerical analysis, cf.\ \cite{Galovashkin2009decomp,Jagtap2020cpinn}. We also show how to combine the domain fragmentation with the newly developed collocation polynomial neural forms. \par 

Let us note that this article is a substantial extension of our conference paper \cite{Schneidereit2021PNF}.
There we briefly introduced just the principle of the collocation-based polynomial NF.
Here we present a much more detailed investigation of the collocation-based NF, and as a substantial novelty the expansion and combination with our novel domain fragmentation method for solving initial value problems. That enables us to discuss both methods in a merged context, which we find highly suitable. Moreover, we updated the notation to a more suitable form.

\paragraph{Related work and problem statement} In our previous work \cite{Schneidereit2020Study} we have shown that even simple feedforward neural networks (5 hidden layer neurons) are capable of solving a stiff initial value problem. The investigated and studied neural forms approaches are based on \cite{lagaris1998artificial,piscopo2019solving}. Please find the additional related work to be addressed in the previous introduction paragraphs. The best results in the above mentioned computational study were provided by random weight initialisation, but with the drawback of a spread between the results with different random initialisation for unchanged computational parameters. There we now want to improve constant weight initialisation. The main advantage of the latter is that results are exactly reproducible with unchanged computational parameters. Based on the neural forms approach \cite{lagaris1998artificial} we have already shown that a polynomial extension leads to a significant increase of numerical accuracy for constant weight initialisation \cite{Schneidereit2021PNF}. In the present work we now combine the polynomial neural forms and employ domain fragmentation to split the solution domain into smaller subdomains. This technique solves the neural forms on such subdomains with the initial values provided by previous subdomains. The inherent flexibility enables the approach to achieve useful results, even on fairly large domains.

\section{Setting up the neural form (NF)}

In this section, we first recall the TSM and its modified version mTSM, respectively, compare \cite{lagaris1998artificial,piscopo2019solving}. Then we proceed with details on the feedforward neural networks we employ, followed by a description of the novel collocation-based neural form and the subdomain approach.

\subsection{Construction of the neural form}

Consider an initial value problem written in a general form as
\begin{equation}
G\left(t,u(t),\dot{u}(t)\right)=0,~~~u(t_0)=u_0,~~~t\in D\subset\mathbb{R}
\label{ivp-1}
\end{equation}
with given initial value $\ds u(t_0)=u_0$. In order to connect $\ds G$ with a neural network, several approaches introduce a NF as a differentiable function $\ds \tilde{u}(t,\vecp)$, where the vector $\ds \vecp$ contains the network weights. With the collocation method we discretise the domain $\ds D$ by a uniform grid with $\ds n+1$ grid points $\ds t_i$ ($\ds t_0<t_1<\ldots<t_n$), so that the initial value problem \eqref{ivp-1} leads to the formulation
\begin{equation}
G\left(t_i,\tilde{u}(t_i,\vecp),\dot{\tilde{u}}(t_i,\vecp)\right)=\mathbf{0}
\label{ivp-2}
\end{equation}
Let us note that, in a slight abuse of notation, we identify
$\ds G\left(t_i,\tilde{u}(t_i,\vecp),\dot{\tilde{u}}(t_i,\vecp)\right)$ 
with the vector of corresponding entries (grid points), since this enables to give many formula a more elegant,
compact notation. \par
In order to satisfy the given initial value, TSM \cite{lagaris1998artificial} employs the NF as a sum of two terms
\begin{equation}
\tilde{u}(t,\vecp)=A(t)+F(t,N(t,\vecp))
\label{TS_TSM}
\end{equation}  
where $\ds A(t)$ is supposed to match the initial condition (with the simplest choice to be $\ds A(t)=u(t_0)$), while $\ds F(t,N(t,\vecp))$ is constructed to eliminate the impact of $\ds N(t,\vecp)$ at $\ds t_0$. The choice of $\ds F(t,N(t,\vecp))$ determines the influence of $\ds N(t,\vecp)$ over the domain. \par 
Since the NF as used in this work satisfies given initial values by construction, we define the corresponding cost function incorporating Eq.\ \eqref{TS_TSM} as
\begin{equation}
E[\vec{p}]=\frac{1}{2}\bigg\lVert G\left(t_i,\tilde{u}(t_i,\vecp),\dot{\tilde{u}}(t_i,\vecp)\right)\bigg\rVert_2^2
\label{costTSM}
\end{equation}
which is subject to minimisation. Although Eq.\ \eqref{ivp-2} denotes a system of equations which may be solvable w.r.t. $\ds \vecp$, the actual equation of interest is Eq.\ \eqref{costTSM} and its optimisation will return suitable neural network weights. \par 
Let us now turn to the mTSM approach after \cite{piscopo2019solving}. The mTSM approach chooses the NF to be equivalent to the neural network output directly
\begin{equation}
\tilde{u}(t,\vecp)=N(t,\vecp)
\label{TS_mTSM}
\end{equation}
Since no condition is imposed by the initial value on the NF in this way, the conditions are added to the cost function when relying on Eq.\ \eqref{TS_mTSM}:
\begin{equation}
E[\vecp]=\frac{1}{2}\bigg\lVert G\left(t_i,\tilde{u}(t_i,\vecp),\dot{\tilde{u}}(t_i,\vecp)\right)\bigg\rVert _2^2+\frac{1}{2}\bigg\lVert N(t_0,\vecp)-u(t_0)\bigg\rVert_2^2
\label{costmTSM}
\end{equation}

\subsection{Neural network architecture and optimisation}

In this section we will describe how a feedforward neural network with one hidden layer operates in our setting. Specific variants will be addressed in the corresponding sections. \par

\begin{wrapfigure}[18]{r}{0.4\textwidth}
\includegraphics[scale=0.75]{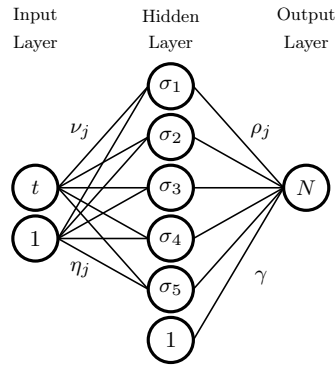}
\caption{\label{ANNarchitecture} The architecture of our incorporated feedforward neural network with one hidden layer bias.}
\end{wrapfigure}

As depicted in Fig.\ \ref{ANNarchitecture} we employ one hidden layer, with $H$ hidden layer neurons supplemented by one bias neuron. Having in addition one bias neuron in the input layer and a linear output layer neuron, the neural network output reads 
\begin{equation}
N(t,\vecp)=\sum_{j=1}^{H}\rho_j\sigma_j+\gamma
\end{equation}
Thereby $\ds \sigma_j=\sigma(z_j)=1/(1+e^{-z_j})$ represents the sigmoid activation function with the weighted sum $\ds z_j=\nu_jt+\eta_j$. Here, $\ds \nu_j$ (input layer neuron), $\ds \eta_j$ (input layer bias neuron), $\ds \rho_j$ (hidden layer neurons) and $\ds \gamma$ (hidden layer bias neuron) denote the weights which are stored in the weight vector $\ds \vecp$. The input layer passes the domain data $t$ (that is in practice $t_i$), weighted by $\nu_j$ and $\eta_j$, to the hidden layer for processing. The neural network output $N(t,\vecp)$ is again a weighted sum of the values $\ds \rho_j \sigma(z_j)$ with $\ds \gamma$ added. With $\ds N(t,\vecp)$ given, the neural forms and cost functions in Eqs.\ \eqref{costTSM},\eqref{costmTSM}, are obtained.\par
As usual, the cost function gradient is used to update $\ds \vecp$ in order to find a (local) minimum in the weight space. One training cycle is called an epoch and consists of a full iteration over all training points points. The gradient, with respect to the network weights, determines their influence on the network output. With this information, each incorporated weight can be adjusted individually to lead the network into a local minimum during the training process. For optimising the cost function we consider here Adam (adaptive moment estimation) \cite{kingma2017Adam} which is a stochastic gradient descent method, using adaptive learning for every weight.\par
If a weight update is performed after the gradient computation for a single grid point we call this method single batch training (SBtraining) here. An alternative proceeding, performing the weight update after a complete iteration over all grid points, averaging the cost function gradient and training error, is denoted here as full batch training (FBtraining). \par
Let us comment in some detail on the relation between grid points and training points. Our setting is an unsupervised learning framework, where grid points are used for domain 
discretisation, and where the unknowns are values of the ODE solution at exactly these grid points. Thus, in our setting, the grid points are identical with the training points.
Let us stress in this context, that the grid points by themselves stay fixed during network optimisation. \par

\section{The novel collocation neural form (CNF)}

Making Eq.\ \eqref{TS_TSM} precise for our grid-based setting, a suitable choice for the neural form of TSM with given $\ds u(0)=u_0$ is
\begin{equation}
\tilde{u}(t,\vecp)=u_0+N(t,\vecp)t
\label{u-t-startpoint}
\end{equation}
where $t$ will be evaluated at grid points $t_i$. Please note, if the initial point is different from $\ds t_0=0$, this results in a shift of $\ds t\to (t-t_0)$ in Eq.\ \eqref{u-t-startpoint}. Compared to a first order polynomial $\ds q_1(t)=a_0+a_1t$ one may find similarities in the structure. Motivated by the expansion of an $m$-th order collocation function polynomial \cite{Antia2012book}
\begin{equation}
q_m(t)=a_0+\sum_{k=1}^ma_kt^k 
\end{equation}
we are lead to set up our collocation-based NF (CNF) approach for TSM:
\begin{equation}
\tilde{u}_C(t,\mat{P}_m)=u_0+\sum_{k=1}^mN_k(t,\vecp_k)t^k
\label{newTS_TSM}
\end{equation}
The weight vector is denoted by $\ds \vecp_k$ and we define the matrix $\ds \mat{P}_m$ of $\ds m$ weight vectors $\ds \mat{P}_m=(\vecp_1,\ldots,\vecp_m)$. \par
The use of higher order monomial powers $\ds t^k$ as in Eq.\ \eqref{newTS_TSM} not only generalises previous methods, but may also enable better stability and accuracy properties, as we show in this paper. Let us also observe, that the neural networks take on the roles of coefficient functions for the values of $\ds t^k$. We conjecture at this point that this construction makes sense since in this way several possible multipliers (not only $\ds t$ as in Eq.\ \eqref{u-t-startpoint}) are included for neural form construction. It is important to mention that the new neural form construction Eq.\ \eqref{newTS_TSM} fulfills the initial condition. \par
Let us stress that the proposed ansatz (Eq.\ \eqref{newTS_TSM}) includes $\ds m$ neural networks, where $\ds N_k(t,\vecp_k)$ represents the $\ds k$-th neural network 
\begin{equation}
N_k(t,\vecp_k)=\sum_{j=1}^H\rho_{j,k}\sigma(\nu_{j,k}t+\eta_{j,k})+\gamma_k 
\end{equation}
The corresponding cost function is then given as in Eq.\ \eqref{costTSM}. \par
We extend the mTSM method in a similar way as we obtained the TSM extension in Eq.\ \eqref{newTS_TSM}:
\begin{equation}
\tilde{u}_C(t,\mat{P}_m)=N_1(t,\vecp_1)+\sum_{k=2}^mN_k(t,\vecp_k)t^{k-1}
\label{newTS_mTSM}
\end{equation}
Thereby the first neural network $\ds N_1(t,\vecp_1)$ is set to learn the initial condition in the same way as stated in Eq.\ \eqref{costmTSM}. \par
From now on we will refer to the number of neural networks in the neural form as the collocation neural form order, denoted by $\ds m$.

\section{The novel subdomain collocation neural form (SCNF)}

The previous described TSM and mTSM approaches use the IVP structure together with the given initial value in order to train the neural networks on a certain domain. In a prior experimental study \cite{Schneidereit2020ODEANN} we figured out that especially TSM tends to struggle with approximating the solution on larger domains. However, on small domains the numerical error tends to remain small. 
Since the domain variable $t$ effectively acts as a scaling of $\ds N(t, \vecp)$, we conjecture that a large domain size variation
may introduce the need for a higher amount of training points or the use of a more complex neural network architecture.
\par 
These circumstances motivate us to introduce a second stage of discretising the domain. That is, we split the solution domain $\ds D$ in subdomains $\ds D_l,l=1,\ldots,h$, with $\ds n+1$ grid points $\ds t_{i,l}$ in each subdomain. Now the CNF is solved separately in each subdomain. The interfacing grid points overlap, i.e.\ the computed value $\ds \tilde{u}_C(t_{n,l-1},\mat{P}_{m,l-1})$ at the last grid point of any subdomain $D_{l-1}$ is set to be the new initial value $\ds \tilde{u}_C(t_{0,l},\mat{P}_{m,l})$ for the next subdomain $\ds D_l$. \par 
Since the CNF for TSM is constructed in order to satisfy the given initial values, we force the subdomain CNF (SCNF) to also hold that characteristic. Therefore the SCNF is constructed to satisfy the new initial values in each subdomain, namely 
\begin{equation}
\tilde{u}_C(t_{i,l},\mat{P}_{m,l})=\tilde{u}_C(t_{0,l},\mat{P}_{m,l})+\sum_{k=1}^mN_k(t_{i,l},\vecp_{k,l})(t_{i,l}-t_{0,l})^k 
\end{equation}
The neural networks are now scaled by $\ds (t_{i,l}-t_{0,l})^k$, which in fact may avoid higher scaling factors, depending on the subdomain size. The arising cost function, similar to Eq.\ \eqref{costTSM}, is
\begin{equation}
E_l[\mat{P}_{m,l}]=\frac{1}{2}\bigg\lVert G\left(t_{i,l},\tilde{u}_C(t_{i,l},\mat{P}_{m,l}),\dot{\tilde{u}}_C(t_{i,l},\mat{P}_{m,l}) \right)\bigg\rVert_2^2
\label{costTSM2}
\end{equation}
Proceeding to mTSM, we also adopt the CNF approach and set the first neural network to learn the new initial values in each subdomain. That is, the SCNF reads 
\begin{equation}
\tilde{u}_C(t_{i,l},\mat{P}_{m,l})=N_1(t_{i,l},\vecp_{1,l})+\sum_{k=2}^mN_k(t_{i,l},\vecp_{k,l})(t_{i,l}-t_{0,l})^{k-1}
\end{equation}
and the corresponding cost function
\begin{align}
\label{costMTSM2}
E_l[\mat{P}_{m,l}]=&\frac{1}{2}\bigg\lVert G\left(t_{i,l},\tilde{u}_C(t_{i,l},\mat{P}_{m,l}),\dot{ \tilde{u}}_C(t_{i,l},\mat{P}_{m,l}) \right)\bigg\rVert_2^2+ \\ \nonumber &\frac{1}{2}\bigg\lVert N(t_{0,l},\vecp_{1,l})-\tilde{u}_C(t_{0,l},\mat{P}_{m,l})\bigg\rVert_2^2 \nonumber
\end{align}
Let us note at this point, that $\ds G$ in Eq.\ \eqref{costTSM2} and \eqref{costMTSM2} shares the same structure as the general problem in Eq.\ \eqref{ivp-1}. However, the original solution function $\ds u(t)$ has been replaced by the SCNF $\ds \tilde{u}_C(t_{i,l},\mat{P}_{m,l})$. Therefore, $\ds G$ involved in the cost function now relies on one or more neural networks, depending on the neural forms order. Once trained, each subdomain has its unique learned weight matrix $\ds \mat{P}_{m,l}$ which can later be used to recreate the solution or evaluate the solution at grid points intermediate to the training points. \par 
In order to keep the overview of all terms and indices, we sum them up again: The $\ds i$-th grid point in the $\ds l$-th subdomain is denoted by $\ds t_{i,l}$, while $\ds t_{0,l}$ is the initial point in the subdomain $\ds D_l$ with the initial value $\ds \tilde{u}_C(t_{0,l},\mat{P}_{m,l})$. That is, $\ds t_{n,l-1}$ and $\ds t_{0,l}$ are overlapping grid points. In $\ds D_1$, $\tilde{u}_C(t_{0,1},\mat{P}_{m,1})=u(t_0)$ holds. The matrix $\ds \mat{P}_{m,l}$ contains the set of the $m$ neural network weights in the corresponding subdomain $\ds l$. Finally, $\ds N_k(t_{i,l},\vecp_{k,l})$ denotes the $\ds k$-th neural network in $\ds D_l$.

\section{Experiments and results}
\label{ExpAndRes}

This section is divided into experiments on the collocation neural form (CNF), followed by experiments on the subdomain collocation neural form (SCNF). Prior to this, we will provide detailed information about how the weight initialisation for the different neural networks are realised. The discussion of constant weight initialisation is also one of the main subjects in the experimental section. As stated before, the specific neural network configurations will be addressed in the subsequent experiments. \par 
Weight initialisation with $\ds \vecpConst$ applies to all corresponding neural networks so that they use the same initial values. Increasing the m for the initialisation with $\ds \vecpRnd$ works systematically. For $\ds m=1$, a set of random weights for the neural network is generated. For $\ds m=2$ (now with two neural networks), the first neural network is again initialised with the generated weights from $m=1$, while for neural network number two, a new set of weights is generated. This holds for all $\ds m$ for higher orders, subsequently, in all experiments. To achieve comparability, the same random initialised weights are used in all experiments. \par 
For optimisation we use Adam, which parameters are fixed with, as employed in \cite{kingma2017Adam}, $\ds \alpha=1$e-3, $\ds \beta_1=9$e-1, $\ds \beta_2=9.99$e-1 and $\ds \epsilon=1$e-8. \par

\subsection{Experiments on the collocation neural form (CNF)}
\label{expCNF}

In this section, we want to test our novel CNF approach with the initial value problem
\begin{equation}
\dot{u}(t)+5u(t)=0,~~~u(0)=1
\label{eqDahlquist}
\end{equation}
which has the analytical solution $\ds u(t)=e^{-5t}$ and is solved over the entire domain $\ds D=[0,2]$ (without domain fragmentation). The Eq.\ \eqref{eqDahlquist} involves a damping mechanism, making this a simple model for stiff phenomena \cite{Dahlquist1978stiff}. \par
The numerical error $\ds \Delta u$ shown in subsequent diagrams in this section is defined as the $\ds l_1$-norm of the difference between the exact solution and the corresponding CNF
\begin{equation}
\Delta u=\frac{1}{n+1}\big\lVert u(t_i)-\tilde{u}_C(t_i,\mat{P}_m)\big\rVert_1 
\end{equation}
If we do not say otherwise, the fixed computational parameters in the subsequent experiments are: 1 input layer bias, 1 hidden layer with 5 sigmoid neurons, 1e5 training epochs, 10 training points, $\ds D=[0,2]$ and the weight initialisation values which are $\ds \vecpConst=-10$ and $\ds \vecpRnd\in[-10.5,-9.5]$. These values may seem arbitrarily chosen, but we found them to work well for both TSM and mTSM so that a useful comparison is possible.

\paragraph{Weight initialisation} 
\begin{wraptable}[11]{r}{0.4\textwidth}
\centering
\vspace{-0.25cm}
\begin{tabular}{c|c|c}
No. & $\ds \Delta u(\vecpRnd)$ & $\ds \Delta u(\vecpConst)$ \\[1ex]
\hline
1 &~5.7148e-6~&~2.6653e-6 \\
2 &~7.5397e-6~&~2.6653e-6 \\
3 &~3.7249e-5~&~2.6653e-6 \\
4 &~1.1894e-5~&~2.6653e-6 \\
5 &~7.7956e-6~&~2.6653e-6 \\
\end{tabular}
\caption{\label{table_results} Results for five different realisations
during optimisation (mTSM, $\ds m=2$)}
\end{wraptable}
Let us comment in some more detail on weight initialisation. The weight initialisation plays an important role and determines the starting point for gradient descent. Poorly chosen, the optimisation method may fail to find a suitable local minimum. The initial neural network weights are commonly chosen as small random values \cite{Fernandez2001Weights}. Let us note that this is sometimes considered as a computational characteristic of the stochastic gradient descent optimisation. Another option is to choose the initialisation to be constant. This method is not commonly used for the optimisation of neural networks since random weight initialisation may lead to better results. However, constant initialisation returns reliably results of reasonable quality if the computational parameters in the network remain unchanged. \par
As previous experiments have documented \cite{Schneidereit2020ODEANN,lagaris1998artificial,piscopo2019solving}, both TSM and mTSM are able to solve differential equations up to a certain degree of accuracy. However, an example illustrating the accuracy of five computations with random weights $\ds \vecpRnd$ respectively constant weights $\ds \vecpConst$ shows that the quality of approximations may vary considerably, see
Table  \ref{table_results}. As observed in many experiments, even a small discrepancy in the initialisation with several sets of random weights in the same range, may lead to a significant difference in accuracy. On the other hand, the network initialisation with constant values very often gives reliable results by the proposed novel approach. This motivates us to study in detail the effects of constant network initialisation.

\subsubsection{CNF Experiment: number of training epochs}
\label{exp5.1.1}

\begin{figure*}[!hb]
    \centering
    \includegraphics{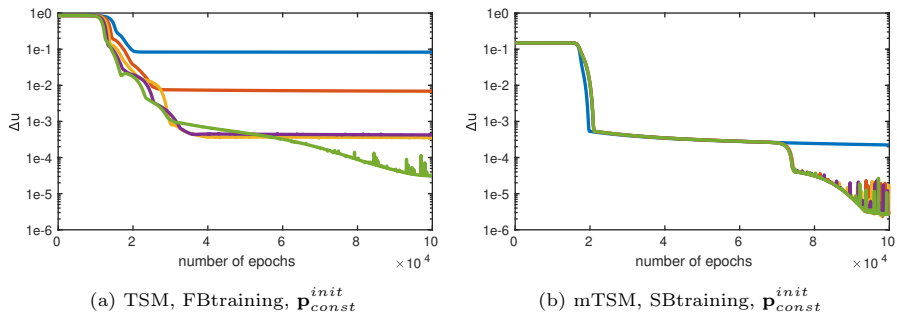}
    \caption{\label{FIGexp5.1.1} {\bf Experiment in \ref{exp5.1.1}} Number of training epochs, (blue) $\ds m=1$, (orange) $\ds m=2$, (yellow) $\ds m=3$, (purple) $\ds m=4$, (green) $\ds m=5$}
\end{figure*}
The first experiment shows for different m how the numerical error $\ds \Delta u$ behaves depending on the number of training epochs. The diagrams only display every hundredth data point. \par
In Fig.\ \ref{FIGexp5.1.1}(a) with TSM and $\ds \vecpConst$ results for $\ds m=1$ (blue) do not provide any useful approximation, independent of the batch training method selected. With a second neural network for $\ds m=2$ (orange) in the neural form, $\ds \Delta u$ approximately lowers by one order of magnitude so that we now obtain a solution which can be considered to rank at the lower end of reliability. However, the most interesting result in Fig.\ \ref{FIGexp5.1.1}(a) is $\ds m=5$ (green) with the best accuracy at the end of the optimisation process but with the drawback of occurring oscillations. These may arise by the chosen optimisation method. \par
Table \ref{TSM_RK4} shows the numeric error $\ds \Delta u(t_i)$ at individual grid points $\ds t_i$. For both CNF and Runge-Kutta 4 (RK4) the results were computed with ten grid points, resulting in the CNF approach with $\ds \vecpConst$ performing better over Runge-Kutta 4. However, further refining the grid for Runge-Kutta 4 will result in significantly lower $\ds \Delta u(t_i)$. \par 
For mTSM with SBtraining and $\ds \vecpConst$, already $\ds m=1$ converges to a solution accuracy that can be considered reliable. However, we observe within Fig.\ \ref{FIGexp5.1.1}(b) that only the transition from $\ds m=1$ (blue) to $\ds m=2$ (orange) affects $\ds \Delta u$ with increasing accuracy, while heavy oscillations start to occur. \par
\begin{wraptable}[16]{r}{0.5\textwidth} 
\centering
\vspace{-0.4cm}
\begin{tabular}{c|c|c}
$\ds t_i$ & $\ds \Delta u(t_i)$ (TSM) & $\ds \Delta u(t_i)$ (RK4) \\[1ex]
\hline
$0.00$ & 0.0000e0 & 0.0000e0 \\
$0.22$ & 3.8158e-5 & 0.1769e0  \\
$0.44$ & 3.5101e-5 & 0.1478e0 \\
$0.66$ & 1.4318e-5 & 9.4013e-2 \\
$0.88$ & 1.2001e-5 & 5.3900e-2 \\
$1.11$ & 4.5407e-5 & 2.9361e-2 \\
$1.33$ & 5.2069e-6 & 1.5546e-2 \\
$1.55$ & 6.6105e-5 & 8.0942e-3 \\
$1.77$ & 1.2052e-5 & 4.1712e-3 \\
$2.00$ & 7.9787e-5 & 2.1357e-3 \\
\end{tabular}
\caption{\label{TSM_RK4} Numerical error comparison at individual grid points with $m=5$ and $\ds \vecpConst$}
\end{wraptable}
In not documented results with $\ds \vecpRnd$, m has only minor influence on the accuracy. Especially FBtraining for mTSM shows the same trend for both initialisation methods with only minor differences in the last epochs. \par  
Let us note that the displayed results show the best approximations using constant or random initialisation. This means, we obtain the best results for TSM with FBtraining, $\ds m=5$ (green) and for mTSM with SBtraining, $\ds m\ge2$, respectively. \par 
Concluding this experiment, we were able to get better results with $\vecpConst$ over $\ds \vecpRnd$. Increasing the m to at least order five seems to be a good option for TSM and FBtraining, whereas further m may provide even better approximations. For mTSM we can not observe benefits for m above order 2. \par 
Moreover, we see especially that the increase in the order of the neural form in \eqref{newTS_TSM} appears to have a similar impact on solution accuracy as the discretisation order in classical numerical analysis.


\subsubsection{CNF Experiment: domain size variation}
\label{exp5.1.2}

\begin{figure*}[!ht]
    \centering
    \includegraphics{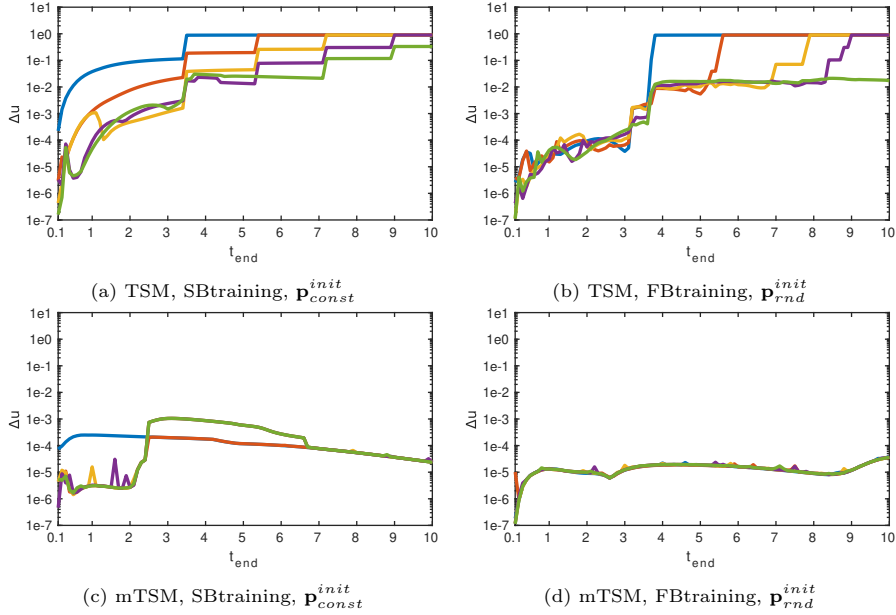}
    \caption{\label{FIGexp5.1.2} {\bf Experiment in \ref{exp5.1.2}} Domain size variation, (blue) $m=1$, (orange) $m=2$, (yellow) $m=3$, (purple) $m=4$, (green) $m=5$}
\end{figure*}
Investigating the methods concerning different domain sizes provides information on the reliability of computations on larger domains. The domains in this experiment read as $\ds D=[0,t_{end}]$ and we directly compare in this experiment $\ds \vecpConst$ with $\ds \vecpRnd$. \par 
In Fig.\ \ref{FIGexp5.1.2}(a), \ref{FIGexp5.1.2}(b), we observe TSM from around $t_{end}=3.5$ to incrementally plateau to unreliable approximations. 
Increasing m improves $\ds \Delta u$ on small domains and shifts the observable step-like accuracy degeneration towards larger domains. \par
However, even with $m=5$ (green) the results starting from domain size $t_{end}=3.5$ towards larger sizes are unreliable. Previous to the first plateau higher m provide significant better $\ds \Delta u$ for $\ds \vecpConst$, while there are only minor changes for $\ds \vecpRnd$ for the TSM method. This holds for both SBtraining and FBtraining, and one can say that in this experiment TSM works better with $\ds \vecpRnd$, even without increasing m. \par
Turning to the mTSM extension, we observe in Fig.\ \ref{FIGexp5.1.2}(c) with SBtraining the existence of a certain point from where different m return equal values, whereas FBtraining returns (close to) equal results for all the investigated domain sizes. However, we see some evidence for the use of $m=2$ (orange) over $\ds m=1$ (blue) to show an overall good performance. A further increase of m is not necessary with this approach, confirming results from Experiment \ref{exp5.1.1}. \par 
Let us also note that, with mTSM we find that a small domain seems to favour $\ds \vecpConst$ which then provides better results than $\ds \vecpRnd$. 

\subsubsection{CNF Experiment: number of training points variation}
\label{exp5.1.3}

\begin{figure*}[!ht]
    \centering
    \includegraphics{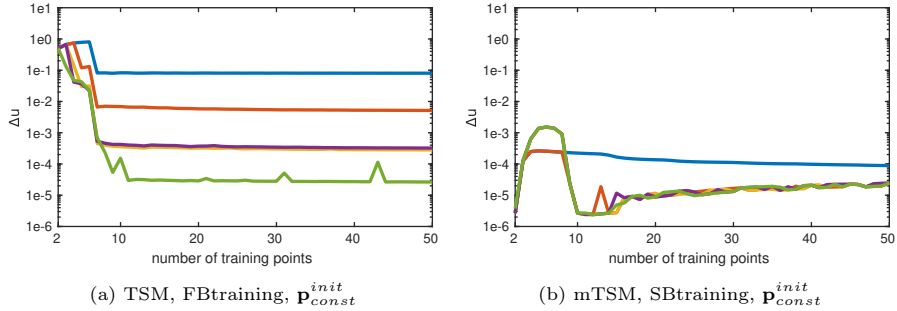}
    \caption{\label{FIGexp5.1.3} {\bf Experiment in \ref{exp5.1.3}} Number of training points variation, (blue) $\ds m=1$, (orange) $\ds m=2$, (yellow) $\ds m=3$, (purple) $\ds m=4$, (green) $\ds m=5$}
\end{figure*}
The behaviour of numerical methods highly depend on the chosen amount of grid points, so that in this experiment we analogously investigate the influence of the number of training points (nTP). In every computation, the domain $\ds D$ is discretised by equidistant grid points. \par
As in the previous experiments, the m shows a major influence on the results with TSM, and the best approximations are provided by $\ds \vecpConst$ with $\ds m=5$ (green) as seen in Fig.\ \ref{FIGexp5.1.3}(a). An interesting behaviour (observed also in a different context in Fig.\ \ref{FIGexp5.1.1}(a)) is the equivalence between $\ds m=3$ (yellow) and $\ds m=4$ (purple). Both converge to almost exactly the same $\ds \Delta u$, where one may assume a saturation for the m. However, another increase in the order decreases the numerical error again by one order of accuracy. \par
Turning to mTSM in Fig. \ref{FIGexp5.1.3}(b) we again find a major increase in accuracy after a transition from $\ds m=1$ (blue) to $\ds m=2$. For nTP $=50$, values for $\ds m\ge 2$ converge to the same results as provided by TSM with $\ds m=5$. \par
Concluding this experiment, we again find evidence that increasing m in the proposed approach provides an improved accuracy for $\ds \vecpConst$. However, increasing nTP seems not to improve the accuracy from a certain point on, unlike for numerical methods. But one could argue, that the analogy between the number of grid points for numerical methods here is the number of epochs. 

\subsection{Experiments on the subdomain collocation neural form (SCNF)}
\label{expSCNF}

\begin{figure}[!h]
\centering
\includegraphics[scale=1.25]{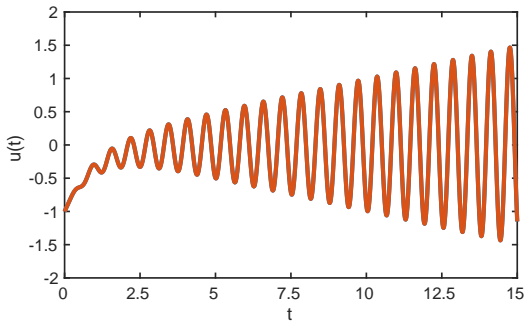}
\caption{\label{IVPsolutionFIG} Analytical solution for initial value problem \eqref{IVP}}
\end{figure}

In Section \ref{expCNF}, while the test equation is stiff, its solution is at the same time very smooth and the equation is solved on a small domain. However, Fig.\ \ref{FIGexp5.1.2} in Experiment \ref{exp5.1.2}, shows that TSM does not provide reliable solutions on larger domains. Hence, we want to show that the novel SCNF approach is able to work even on a fairly large domain with a different initial value problem. Therefore we use the following test equation
\begin{equation}
\dot{u}(t)-t\sin(10t)+u(t)=0,~~~u(0)=-1
\label{IVP}
\end{equation}
with the analytical solution
\begin{equation}
u(t)=\sin(10t)\bigg(\frac{99}{10201}+\frac{t}{101}\bigg)+\cos(10t)\bigg(\frac{20}{10201}-\frac{10t}{101}\bigg)-\frac{10221}{10201}e^{-t}
\label{IVPsolution}
\end{equation}
The solution is shown in Fig.\ \ref{IVPsolutionFIG} for $t\in[0,15]$
and incorporates heavily oscillating and increasing characteristics, similar to instabilities. Although our approach is not limited to certain types of IVPs, we find Eq.\ \eqref{IVP} to represent possible real-world behaviour and find it suitable to serve as an example IVP.  \par  
The numerical error $\ds \Delta u_l$ is now defined as the averaged $\ds l_1$-norm of the difference between the exact solution and the corresponding SCNF in each subdomain
\begin{equation}
\Delta u_l=\frac{1}{n+1}\big\lVert u(t_{i,l})-\tilde{u}_C(t_{i,l},\mat{P}_{m,l})\big\rVert_1 
\end{equation}
whereas $\ds \Delta u$ averages the numerical error of the subdomains
\begin{equation}
\Delta u=\frac{1}{h}\sum_{l=1}^h\Delta u_l
\end{equation}
The weight initialisation works as employed in Section \ref{ExpAndRes} and the values are fixed to $\ds \vecpConst=0$ and $\ds \vecpRnd\in[-0.5,0.5]$. In the subsequent experiments, the solution domain is kept constant to $\ds D=[0,15]$ and the neural networks are training with 1e5 epochs. \par 
In addition we use the method of training the neural networks incrementally which has been employed in \cite{piscopo2019solving}. That is, we initially train the neural networks for the first grid point, afterwards for the first two grid points. We continue the procedure up to a FBtraining of all grid points in each subdomain. The initial weight initialisation is the same in each subdomain.\par 
Please note at this point, that we provide an explicit comparison to the Runge-Kutta 4 method only in the last experiment of this section. We find that our approach provides by construction a very high flexibility, to deal with many types of initial value problems that may require specific constructions in classic numerical analysis (e.g. by symplectic integration). However, in simple settings we will usually find the network based approach at the moment not competitive to numerical state-of-the-art solvers (w.r.t. computational efficiency) which have been developed and refined over decades. We think that because of the much higher flexibility of the network based tool, this comparison would not be entirely adequate. \par 
As an example for the flexibility, a recent neural network approach \cite{Flamant2020bundles} makes it superficial to restart the computation of numerical solutions when considering multiple, different initial conditions. We also demonstrate the flexibility in this paper in Section\ \ref{exp5.2.5} and show how invariants can be simply added to the cost function. Regarding numerical methods for handling those problems, special constructions are often needed.
\par 
In our test example in Eq.\ \eqref{IVP}, we find a graphical comparison in the subsequent experiments to not provide further information since the visual differences between analytical solution and solution with Runge-Kutta-4 with adaptive time stepping are minor.

\paragraph{A scaling experiment}
The original TSM neural form (Eq.\ \eqref{TS_TSM}) is theoretically capable of approximating every continuous function, according to the universal approximation theorem \cite{Cybenko1989uat}. However, Table \ref{table_results2} shows results for a TSM neural form with a single neural network. For different domains we scaled the number of hidden layer neurons linearly and averaged ten computations for each domain with the same computational parameters. \par
\begin{wraptable}[8]{r}{0.5\textwidth} 
\centering
\vspace{-0.25cm}
\begin{tabular}{c|c|c|c}
domain D & $\ds \Delta u$ & Neurons & nTP \\[1ex]
\hline
$[0,1]$ & 8.4228e-4 & 5 & 10 \\
$[0,2]$ & 9.2191e-4 & 10 & 20 \\
$[0,3]$ & 1.9448e-3 & 15 & 30 \\
$[0,4]$ & 1.6751e-2 & 20 & 40 \\
\end{tabular}
\caption{\label{table_results2} TSM neural form, $\ds \vecpRnd$}
\end{wraptable}
The results in Table \ref{table_results2} provide the following message. Increasing the domain size forces the neural network to incorporate more hidden layer neurons and grid points. Indeed, to reach e.g. (averaged) $\ds \Delta u=9.5873$e-4 for $\ds D=[0,3]$, learning the neural network required 50 hidden layer neurons and 75 grid points. In general, determining a suitable architecture in terms of the number of hidden layer neurons and training points is a challenging task. \par 
In subsequent experiments we find the SCNF to be able to solve the initial value problem with neural networks including a small fixed amount of hidden layer neurons and training points in each subdomain. At the same time, this allows to define various important parameters in a simple and straightforward way. 

\subsubsection{SCNF Experiment: CNF versus SCNF}
\label{exp5.2.1}

\begin{figure*}[!ht]
    \centering
    \includegraphics{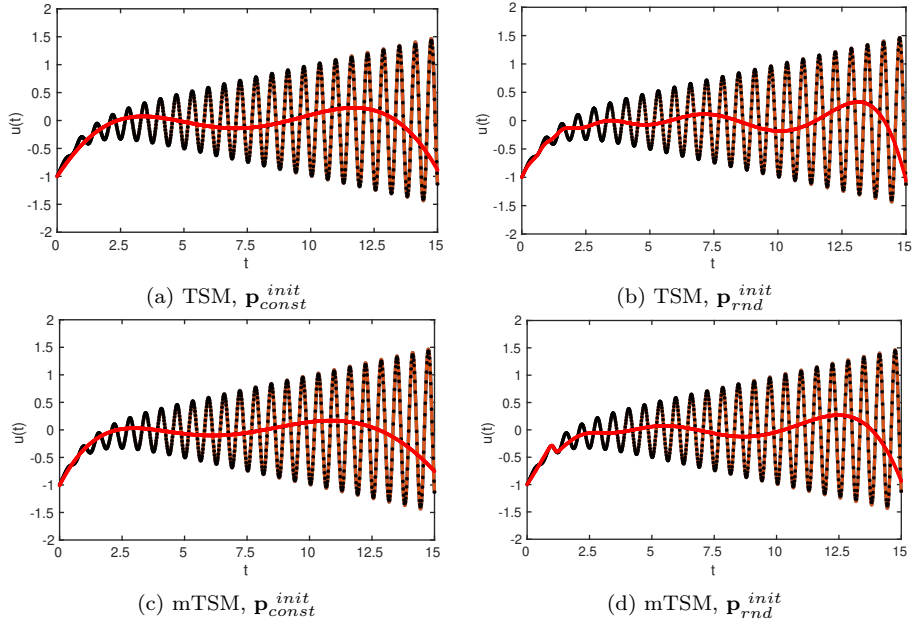}
    \caption{\label{FIGexp5.2.1} {\bf Experiment in \ref{exp5.2.1}} CNF versus SCNF, (orange) analytical solution, (red) CNF solution, (black/dotted) SCNF solution}
\end{figure*}
In the first experiment we compare results of a SCNF with a CNF that is solved over the entire domain. For comparability the total number of training points is constant, namely nTP $=1000$ for the red line and nTP $=10$ with 100 subdomains for the black/dotted line. However, the comparison of two CNFs with the same architecture would not be meaningful because the domain size has a significant influence. Therefore we decided to realise the CNF (red) with a neural network incorporating 1 input layer bias and 100 sigmoid neurons with 1 bias. The SCNF (black/dotted) features neural networks with 1 input layer bias and 5 sigmoid neurons with 1 bias per subdomain. Both CNF and SCNF incorporate $\ds m=3$. In addition we did not increase the domain size incrementally for this experiment, to reduce the number of parameters that prevent comparability. \par
The CNF solution (red) shows throughout all experiments in Fig.\ \ref{FIGexp5.2.1} no useful approximation. In total, the number of hidden layer neurons and training points that would be needed to obtain a useful approximation seems to be much higher. Nonetheless, the SCNF approach (black/dotted) working with the same number of training points was able to solve the initial value problem in a satisfactory way. From a qualitative perspective both TSM and mTSM together with $\ds \vecpConst$ and $\ds \vecpRnd$ provide similar results. \par 
Concluding this experiment, we see that the SCNF method provides a useful solution to the initial value problem. In addition, the incorporated small number of hidden layer neurons enables a much more effective training of the neural networks.

\subsubsection{SCNF Experiment: variation of the neural forms order (m)}
\label{exp5.2.2}

\begin{figure*}[!ht]
\centering
\includegraphics{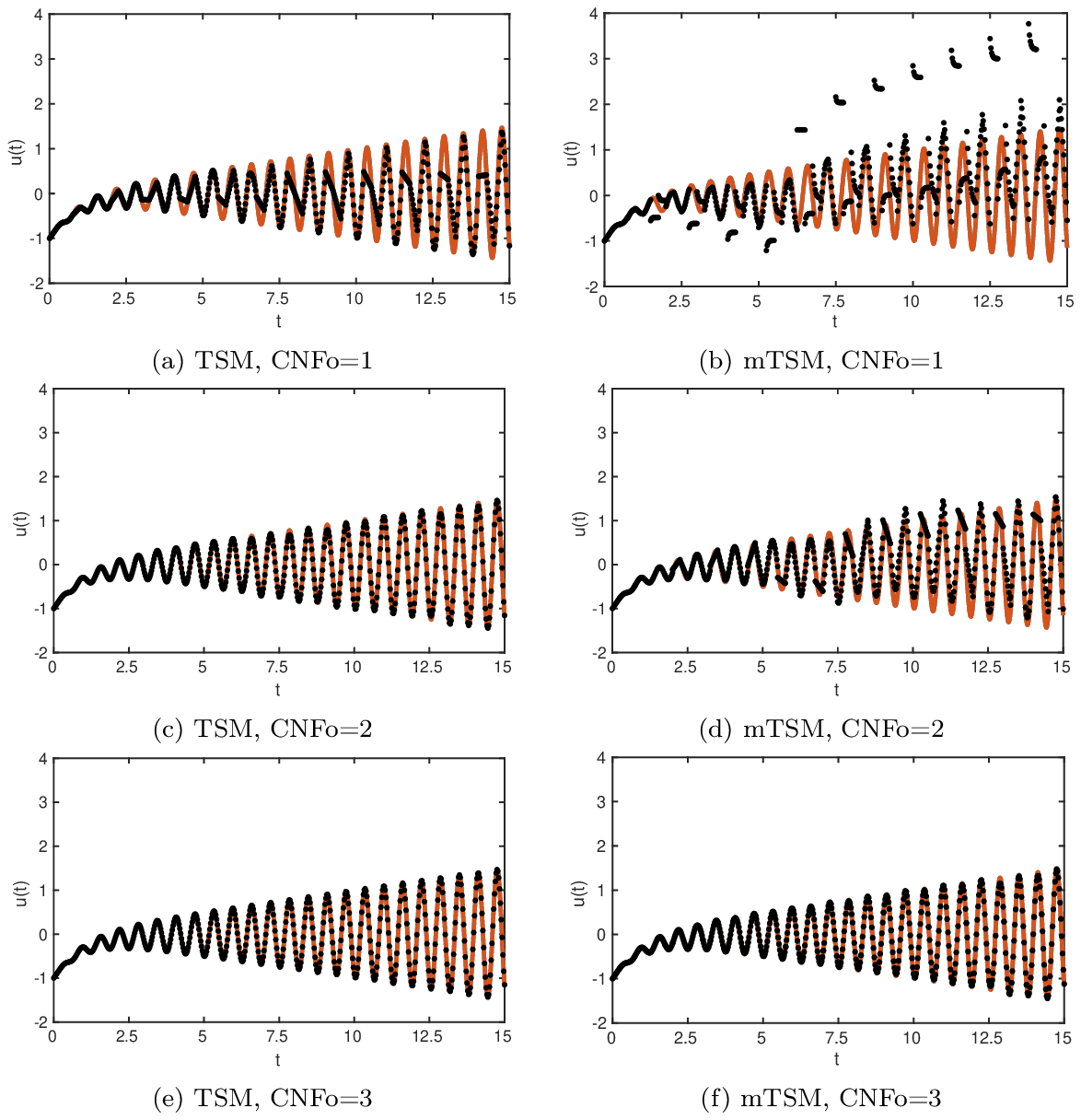}
\caption{\label{FIGexp5.2.2} {\bf Experiment \ref{exp5.2.2}} m variation, $\ds \vecpConst$, (orange) analytical solution, (black/dotted) SCNF solution}
\end{figure*}
\begin{wraptable}[8]{R}{0.5\textwidth}
\centering
\begin{tabular}{c|c|c}
m & $\ds t_{n,l-1}$ & $\ds t_{0,l}$ \\[1ex]
\hline
1 & 2.7834 & 4.8758  \\
2 & 0.5763 & 0.7478  \\
3 & 0.0846 & 0.0848 \\
\end{tabular}
\caption{\label{table_results3} $\ds l_{\infty}$-norm for mTSM interface grid points, $\ds \vecpConst$}
\end{wraptable}
The ability to approximate the initial value problem with SCNF, depending on different m, is subject to this experiment. Here the SCNFs include 1 input layer bias and 5 sigmoid neurons with 1 bias. The solution domain is split into 60 subdomains with 10 grid points in each subdomain. Here, we employ incremental learning in the subdomains. \par 
Results for TSM with $\ds m=1$ in Fig.\ \ref{FIGexp5.2.2}(a) and mTSM with $\ds m=1$ in Fig.\ \ref{FIGexp5.2.2}(b) indicate that the original TSM and mTSM methods are not useful over larger domains, even when employing domain fragmentation. However, the SCNF of first order is able to get back on the solution trend, although several subdomains do not provide correct approximations. In total, both solutions for $\ds m=1$ (especially mTSM) cannot be considered to be reliable.  \par 
That changes for $\ds m=2$, at least for TSM in Fig.\ \ref{FIGexp5.2.2}(c). Here we find, with the exception of some local extreme points, the SCNF to be a reasonable approximation of the initial value problem. This statement however, does not hold for mTSM. Although the general trend now is much closer to the analytical solution, there are still subdomains which do not approximate the solution well. \par  
Results shown in Table \ref{table_results3} represent the $\ds l_{\infty}$-norm of differences between analytical and computed solution, for mTSM as displayed Fig.\ \ref{FIGexp5.2.2}, measured at the last grid points in $\ds D_{l-1}$, namely $\ds t_{n,l-1}$, and the corresponding initial points in $\ds D_l$, $\ds t_{0,l}$. We propose to consider this measure, since it indicates how well the solution can be met over the subdomains. We find that increasing m has a major influence on the accuracy. \par
We conjecture that learning the subdomain initial values becomes easier for mTSM, the more neural networks are incorporated. That is mainly because the first neural network can so to say focus on learning the initial values, while the other networks are more engaged with the IVP structure. We think that this conjecture can be confirmed by the decreasing discrepancy between the overlapping at the interfaces for higher orders of m.  \par  
The overall best solutions here are provided by $\ds m=3$ (Fig.\ \ref{FIGexp5.2.2}(e),\ref{FIGexp5.2.2}(f)) for both TSM and mTSM in this experiment.  \par 
We tend to favor TSM over mTSM, since the initial value in each subdomain is satisfied by the corresponding SCNF (where the learned value at $\ds t_{n,l-1}$ is set to be the initial value for $\ds t_{0,l}$) and does not have to be learned again.

\subsubsection{SCNF Experiment: number of subdomain variation}
\label{exp5.2.3}

\begin{figure*}[!ht]
    \centering
    \includegraphics{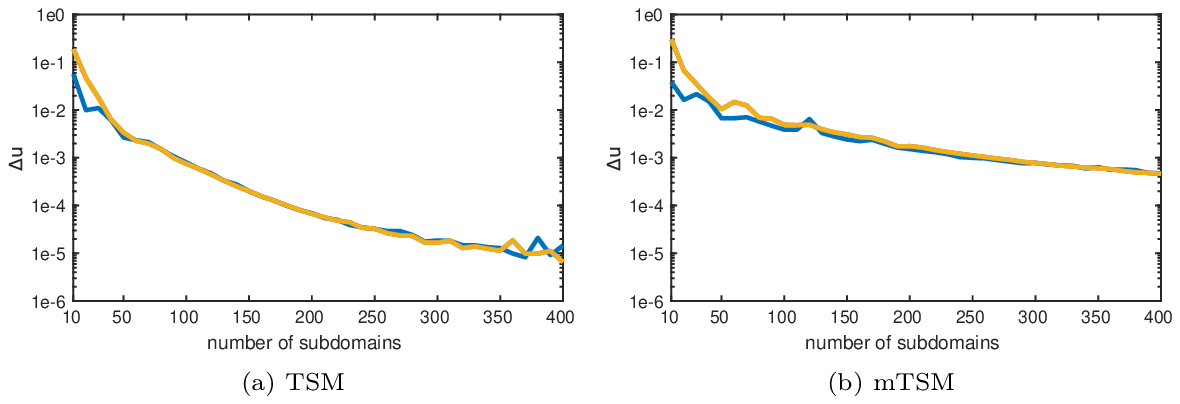}
    \caption{\label{FIGexp5.2.3} {\bf Experiment in \ref{exp5.2.3}} Number of subdomain variation, (blue) $\ds \vecpRnd$, (yellow) $\ds \vecpConst$}
\end{figure*}
In this experiment we investigate the influence of the total number of subdomains on the numeric error $\ds \Delta u$. Fig.\ \ref{FIGexp5.2.3} shows the behaviour for $\ds \vecpRnd$ (blue) and $\ds \vecpConst$ (yellow). The SCNF incorporate $\ds m=3$, 1 input layer bias, 5 sigmoid neurons with 1 bias and 10 grid points in each subdomain. We again employ incremental learning in the subdomains. \par 
Let us first comment on the SCNF for TSM in Fig.\ \ref{FIGexp5.2.3}(a). Despite minor differences between the solutions corresponding to $\ds \vecpRnd$ and $\ds \vecpConst$ for smaller numbers of subdomains, both initialisation methods show a very similar trend. A saturation regime seems to appear for around 350 subdomains with $\ds \Delta u\approx 1$e-5. \par 
Turning to mTSM in Fig.\ \ref{FIGexp5.2.3}(b), we again observe a similar behaviour between the methods with $\ds \vecpRnd$ and $\ds \vecpConst$. Although the differences disappear not before larger numbers of subdomains. We find that even at 400 subdomains the numerical error $\ds \Delta u$ can not compete with TSM here. \par 
Let us note again, that the chosen weight initialisation approach for $\ds \vecpRnd$ (see Section \ref{ExpAndRes}) means that the random weights are initialised in the same way in each subdomain. In undocumented tests we observed that the results may show slight to significant variations, when the random weights are generated independently for each network over the subdomains. However, the results we have shown here using $\ds \vecpRnd$ represent a rather typical trend observed in the results. \par 
In conclusion, one can obtain very good approximations with the TSM SCNF approach for both weight initialisation methods. That means, choosing $\ds \vecpConst$ over $\ds \vecpRnd$ has no downsides, which leads us to again support the use of constant weight initialisation.     

\subsubsection{SCNF Experiment: numerical error in the subdomains}
\label{exp5.2.4}

\begin{figure*}[!ht]
    \centering
    \includegraphics{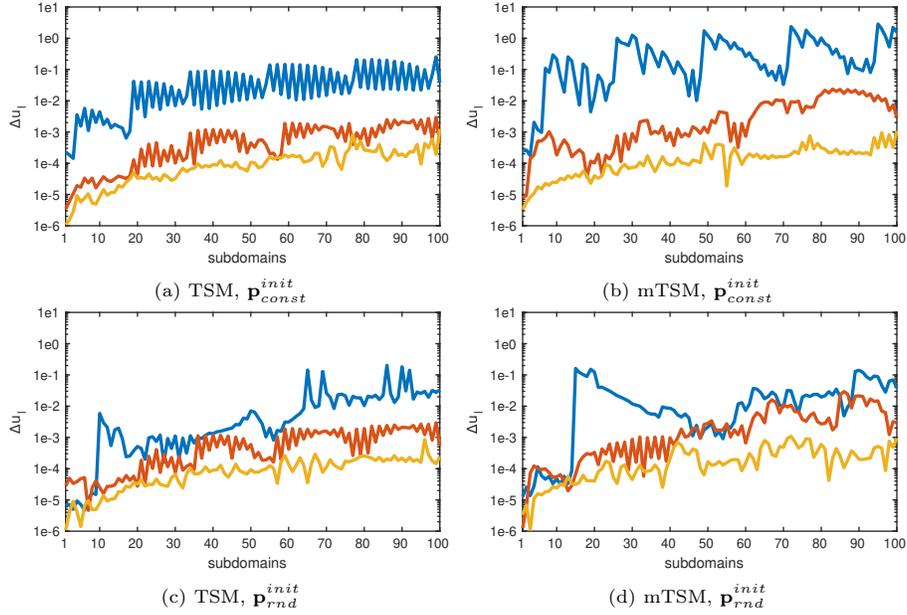}
    \caption{\label{FIGexp5.2.4} {\bf Experiment in \ref{exp5.2.4}} Numerical error in the subdomains, (blue) $\ds m=1$, (orange) $\ds m=3$, (yellow) $\ds m=5$}
\end{figure*}
The last experiment investigates the numeric error $\Delta u_l$ in each subdomain $D_l$, depending on different m. Again, the SCNFs feature 1 input layer bias and 5 sigmoid neurons with 1 bias. The solution is computed with 100 subdomains together with 10 grid points each and incremental learning in the subdomains. \par
Throughout Fig.\ \ref{FIGexp5.2.4}, $\ds m=1$ shows the least good results. Although, if we compare mTSM with $\ds \vecpConst$ in Fig.\ \ref{FIGexp5.2.4}(b) to results for 60 subdomains in Fig.\ \ref{FIGexp5.2.2}(b), increasing the domain fragments by 40 subdomains seems to prevent the solution from diverging. Random weight initialisation works better for $\ds m=1$, especially with TSM. \par   
Solutions provided by $\ds m=3$ and $\ds m=5$ are much better than for $\ds m=1$, and increasing the order clearly tends to increase the accuracy. For TSM with both $\ds m=3$ and $\ds m=5$, as well as for mTSM with $\ds m=5$ from a certain subdomain on, the numerical error saturates. Let us note, that for both $\ds \vecpConst$ and $\ds \vecpRnd$ the differences in the overall numerical error $\ds \Delta u$ are not significant in these cases. \par
In this experiment, we again tend to favour TSM with $\ds \vecpConst$. Although $\ds m=1$ does not work well, the other shown higher orders provide good approximations with saturation regimes. The results confirm our preference of constant initialisation, because $\ds \vecpConst$ does not depend on a good generation of random weights by chance.

\subsubsection{SCNF Experiment: system of initial value problems}
\label{exp5.2.5}

\begin{figure}[!h]
    \centering
    \includegraphics{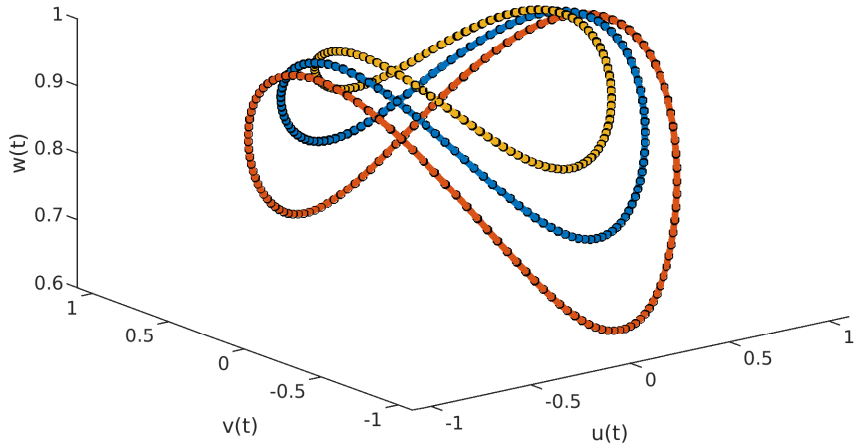}
    \caption{\label{FIGexp5.2.5} {\bf Experiment in \ref{exp5.2.5}} Runge-Kutta 4 (coloured/solid) and SCNF (black/dots) solution  of the free rigid body problem in Eq.\ \eqref{rigid} for different initial values, (blue) \{$u(0)$=$\cos(1.1)$; $v(0)$=$0.6$; $w(0)$=$\sin(1.1)$\}, (orange) \{$u(0)$=$\cos(1)$; $v(0)$=$0.7$; $w(0)$=$\sin(1)$\}, (yellow) \{$u(0)$=$\cos(1.2)$; $v(0)$=$0.5$; $w(0)$=$\sin(1.2)$\}}
\end{figure}

In this section we study a non-linear system of initial value problems. The example system we consider reads
\begin{align}
\label{rigid}
\dot{u}(t)&=a_uv(t)w(t) \\
\dot{v}(t)&=a_vw(t)u(t) \\
\dot{w}(t)&=a_wu(t)v(t) 
\end{align}
with
\begin{equation}
a_u=\frac{I_v-I_w}{I_vI_w},~~~a_v=\frac{I_w-I_u}{I_wI_u},~~~a_w=\frac{I_u-I_v}{I_uI_v}    
\end{equation}
where $\ds I_u,I_v,I_w$ are non-zero real numbers, and given initial values for $\ds u,v,w$. The equations describe the angular momentum of a free rigid body \cite{Griffiths2010NumMethods,Hairer2006GeoNum} with the centre of mass at the origin. These coupled initial value problems are often denoted as Euler equations and feature time invariant characteristics since the independent variable $\ds t$ (time) does not explicitly appear on the right-hand side. The quadratic invariant expression
\begin{equation}
R^2=u^2(t)+v^2(t)+w^2(t)
\label{invariant1}
\end{equation}
conserves the so-called magnitude and describes a sphere, while another quadratic invariant
\begin{equation}
H=\frac{1}{2}\bigg(\frac{u^2(t)}{I_u}+\frac{v^2(t)}{I_v}+\frac{w^2(t)}{I_w}\bigg)
\label{invariant2}
\end{equation}
conserves the kinetic energy and describes an ellipsoid. Both invariant quantities force the solution to stay on the intersection formed by the sphere and the ellipsoid. Since Eqs.\ \eqref{invariant1} and \eqref{invariant2} remain unchanged over time along the solutions, we have
\begin{align}
H&=\frac{1}{2}\bigg(\frac{u^2(0)}{I_u}+\frac{v^2(0)}{I_v}+\frac{w^2(0)}{I_w}\bigg) \\
R^2&=u^2(0)+v^2(0)+w^2(0)
\end{align}
The corresponding initial values $\ds u(0), v(0), w(0)$ are given as in Fig.\ \ref{FIGexp5.2.5} and the fixed principle moments of interior (see \cite{Griffiths2010NumMethods}, Section 14.3) have the values
\begin{equation}
I_u=2,~~~I_v=1,~~~I_w=\frac{2}{3}
\end{equation}
assigned. In general, the neural network approach allows given invariant expressions to be directly added to the cost function, due to its flexibility. For systems of initial value problems, the cost function can be obtained by assigning each solution function its own SCNF and sum up the retrieved $\ds l_2$-norms, here together with the invariant quantities:
\begin{align}
\label{costFctSys}
E[\vec{p}]=\frac{1}{2}&\bigg\lVert
\dot{\tilde{u}}_C-a_u\tilde{v}_C\tilde{w}_C\bigg\rVert_2^2+\frac{1}{2}\bigg\lVert\dot{\tilde{v}}_C-a_v\tilde{w}_C\tilde{u}_C\bigg\rVert_2^2+\frac{1}{2}\bigg\lVert\dot{\tilde{w}}_C-a_w\tilde{u}_C\tilde{v}_C\bigg\rVert_2^2+\\
\frac{1}{2}&\bigg\lVert\tilde{u}_C^2+\tilde{v}^2_C+\tilde{w}^2_C-R^2\bigg\rVert_2^2+\frac{1}{2}\bigg\lVert\frac{\tilde{u}_C^2}{2I_u}+\frac{\tilde{v}^2_C}{2I_v}+\frac{\tilde{w}^2_C}{2I_w}-H\bigg\rVert_2^2 \nonumber
\end{align}
In order to visualise the invariant behaviour, the computed results in Fig.\ \ref{FIGexp5.2.5} are obtained for $\ds t\in[0,30]$, which allows the solution to pass its own initial points more than once. The solution domain is fragmented into 40 subdomains with ten training points in each subdomain. Due to overlapping grid points at the subdomain intersections, the total number of unique training points is nTP $=361$, the same amount was used to obtain computational result with Runge-Kutta 4. Each SCNF has $\ds m=3$ and therefore features three neural networks involved. The latter are initialised with zeros and other training parameters and methods remain unchanged (see Section \ref{expSCNF}). \par 
In Fig.\ \ref{FIGexp5.2.5}, we display both the Runge-Kutta 4 (coloured/solid) and the SCNF (black/dots) solution. As mentioned above, the curves lay on intersections formed between a sphere and an ellipsoid. Changing the point of view in Fig.\ \ref{FIGexp5.2.5} reveals the contours of intersections between the corresponding spheres and ellipsoids for the different initial conditions. \par
\begin{figure}[!h]
  \centering
  \includegraphics[scale=0.75]{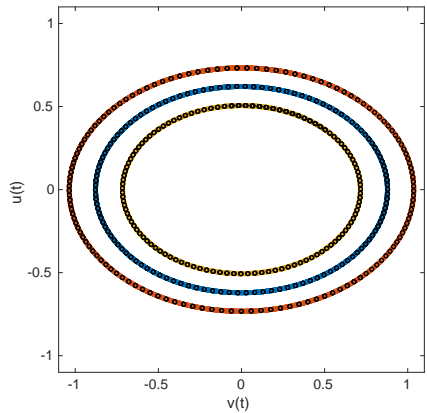}
  \caption{\label{FIGexp5.2.5.2} {\bf Experiment in \ref{exp5.2.5}} Top view of Fig.\ \ref{FIGexp5.2.5} to visualise the spherical and ellipsoidal intersections, colours are adopted from Fig.\ \ref{FIGexp5.2.5}}
\end{figure}
\begin{figure}
\centering
\includegraphics[scale=0.6]{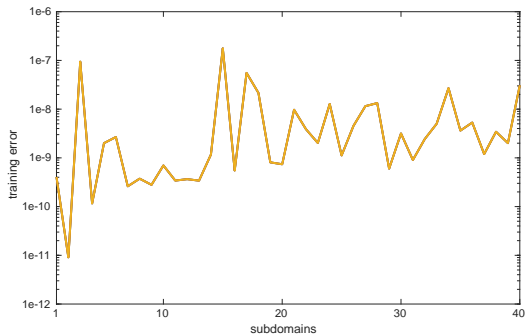}
\caption{\label{FIGexp5.2.5.3} {\bf Experiment in \ref{exp5.2.5}} Training error over the subdomains for \{$u(0)$=$\cos(1.2)$; $v(0)$=$0.5$;$w(0)$=$\sin(1.2)$\}}
\end{figure}
Fig.\ \ref{FIGexp5.2.5.3} shows the training error arising from the cost function in Eq.\ \eqref{costFctSys} over the incorporated subdomains. Since the SCNF solution in each subdomain is computed independently (except for the provided initial value), the training error can differ significantly, even between two adjacent subdomains. Additionally, the flexibility of those independent computations enables this decreasing behaviour of the training error, although the previous subdomain may have shown large errors. \par
In conclusion, the SCNF approach can be easily extended to handle systems of initial value problems, even with nonlinear and time invariant characteristics. A qualitative comparison between the SCNF solution and the Runge-Kutta 4 solution shows only very minor visual differences. As mentioned in Section \ref{expSCNF}, a quantitative comparison may favour Runge-Kutta 4 in terms of the numerical error in this experiment. However, the Runge-Kutta 4 method is not a symplectic integrator and may not preserve quadratic invariants over time.

\section{Conclusion and future work}

The proposed CNF and SCNF approaches merging collocation polynomial basis functions with neural networks and domain fragmentation show clear benefits over the previous neural form constructions. We have studied in detail the constant weight initialisation for our novel CNF approach with a basic stiff initial value problem. Depending on the batch learning methods, the collocation-based extension seems to have some benefits for both TSM and mTSM. For the TSM CNF, this effect is more significant than observed for the mTSM extension. \par 
Focusing on mTSM and the CNF approach, using two neural networks, one for learning the initial value and one multiplied by $t$, seems to have some advantages over other possible mTSM settings. Considering approximation quality as most imperative, we find mTSM with $\ds m=2$ to provide the overall best results for the investigated initial value problem. \par
We find that the proposed SCNF approach combines many advantages of the new developments. Employing higher order CNF methods, it is possible to solve initial value problems over large domains with very high accuracy, and at the same time with reasonable optimisation effort. Moreover, many computational parameters can be fixed easily for this setting, which is a significant issue with other TSM and mTSM variations. \par 
As another important conclusion, in the experiments we were able to show that we can favour constant weight initialisation over random weight initialisation. Nonetheless, the complexity of the IVP has an impact on the architecture and on the values of the initial constant weights. As we have pointed out in a computational study \cite{Schneidereit2020Study}, DE and network related parameters may not be independent of each other. However, the underlying relation between problem complexity and necessary neural network architecture is yet part of future work. \par 
When focusing on constant weight initialisation, we find a further investigation on how to find suitable (constant) initial weights to be of interest. The same holds for the sensitivity of the neural network parameters.\par 
Future research may also include work on other possible collocation functions and on combining the networks with other discretisation methods. In addition to that, let us note that we find optimal control problems to be a possible and relevant potential field of applications for our method, see for instance \cite{Wozniak2017neuro} for recent research in that area.

\section*{Acknowledgement}
This publication was funded by the Graduate Research School (GRS) of the Brandenburg University of Technology Cottbus-Senftenberg. This work is part of the Research Cluster Cognitive Dependable Cyber Physical Systems.

\end{document}